# Mind the Gap: Evaluating the Representativeness of Quantitative Medical Language Reasoning LLM Benchmarks for African Disease Burdens

Authors: Fred Mutisya (MBChB)[1,2], Shikoh Gitau(PhD)[1], Christine Syovata (MBChB)[2], Diana Oigara (MBChB)[2], Ibrahim Matende (MMED)[2], Muna Aden( MBChB)[2], Munira Ali (MD)[2], Ryan Nyotu (MBChB)[2], Diana Marion(MBChB, MBA)[2], Job Nyangena( MBChB)[2], Nasubo Ongoma[1], Keith Mbae[1], Elizabeth Wamicha(PhD)[1], Eric Mibuari ( PhD)[1]  Jean Philbert Nsengemana(MSc, MBA, MPA)[3], Talkmore Chidede (PhD)[4]

Affiliation 1. Qhala, 2. Kenya Medical Association , 3. Africa CDC 4. AfCFTA

## Abstract

**Introduction:** Existing medical LLM benchmarks largely reflect examination syllabi and disease profiles from high-income settings, raising questions about their validity for African deployment where malaria, HIV, TB, sickle-cell disease and other neglected tropical diseases (NTDs) dominate burden and national guidelines drive care. **Methodology:** We systematically reviewed 31 quantitative LLM evaluation papers (Jan 2019–May 2025) identifying 19 English medical QA benchmarks. Alama Health QA was developed using a retrieval augmented generation framework anchored on the Kenyan Clinical Practice Guidelines. Six widely used sets (AfriMed-QA, MMLU-Medical, PubMedQA, MedMCQA, MedQA-USMLE, and guideline-grounded Alama-Health-QA) underwent harmonized semantic profiling (NTD proportion, recency, readability, lexical diversity metrics) and blinded expert rating across five dimensions: clinical relevance, guideline alignment, clarity, distractor plausibility, and language/cultural fit. **Results:** Alama Health-QA captured >40% of all NTD mentions across corpora and the highest within-set frequencies for malaria (7.7%), HIV (4.1%), and TB (5.2%); AfriMed-QA ranked second but lacked formal guideline linkage. Global benchmarks showed minimal representation (e.g.,sickle-cell disease absent in three sets) despite large scale. Qualitatively, Alama scored highest for relevance and guideline alignment; PubMedQA lowest for clinical utility. **Discussion:** Quantitative medical LLM benchmarks widely used in the literature underrepresent African disease burdens and regulatory contexts, risking misleading performance claims. Guideline-anchored, regionally curated resources such as Alama Health-QA—and expanded disease-specific derivatives—are essential for safe, equitable model evaluation and deployment across African health systems.

# 1    Introduction

Large language models (LLMs) now power chatbots, reference tools, and decision-support systems across medicine(Elhaddad & Hamam, 2024). There has been an exponential increase in LLM evaluation research, from fewer than five in 2020 to over thirty in 2023 (Chang et al., 2024). LLMs in medicine are typically evaluated on two fronts: quantitative performance benchmarks (using structured question-answering datasets that allow numeric scoring) and qualitative safety/ethical benchmarks (assessing harms, bias, and alignment). Since 2019, various benchmarks have emerged to measure the capabilities of medical LLMs in both areas, reflecting the growing sophistication and risk awareness in this domain. Yet the datasets used to train and test these models were almost entirely devised for the licensing exams, research priorities, and clinical realities of high-income regions. A survey of these benchmarks by Alaa in 2024 looked at 100 of the most cited medical LLM papers and analyzed the benchmarks used. Sixty percent of the benchmarks used medical school exams while the other 40% mainly used closed-source hospital data (Alaa et al., 2025). For African deployments—where disease burdens, resource constraints, and national treatment protocols differ markedly—this raises two intertwined risks:

1. **Under-representation**: LLM benchmarks share a key bias with LLM training data which is a geographical and demographic bias towards more data rich regions(Algaba et al., 2024). Underrepresentation may span multiple domains including language, disease patterns, demographic patterns etc. Models may misclassify or hallucinate on conditions that dominate African case-mixes (e.g., malaria in pregnancy, sickle-cell crises, severe malnutrition).

2. **Regulatory non-alignment:** A model can score highly on global benchmarks while violating Kenyan, Nigerian, or South-African guidelines—undermining patient safety and regulatory acceptance(Azamfirei et al., 2023).

**Types of Medical LLM Benchmarks**

Benchmarking in medical LLMs can be largely divided into 4 groups: medical language reasoning, medical language generation, medical language understanding and medical summarization(figure 1).

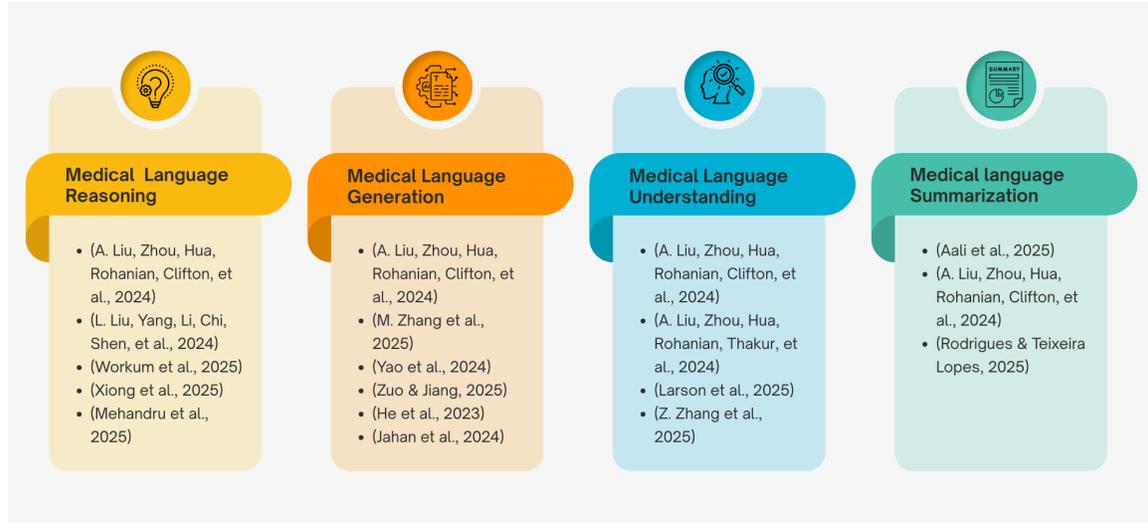

*Figure 1: Types of Medical LLM benchmarks*

Some are standalone benchmarks like the hospital record summarization benchmark with datasets like MIMIC-IV_BHC(Aali et al., 2025). Others like the BenchHealth combine 7 medical tasks like record summarization and 13 representative datasets to make a comprehensive benchmark for medical reasoning, generation and understanding(Liu et al., 2024). BenchHealth is not to be confused with HealthBench, an open source benchmark from OpenAI with over 5000 multiturn conversations between(Arora et al., 2024). Healthbench is one of the most inclusive benchmarking initiatives to date with over 262 physicians from 60 different countries.

Benchmarking medical large language models (LLMs) involves a multi-stage process of dataset design, metric selection, and model evaluation, typically grounded in realistic clinical scenarios or standardized assessments. The reviewed literature demonstrates that many benchmarks are constructed by curating clinically relevant question-answer pairs—often from medical licensing exams, electronic health records (EHRs), or published biomedical texts. For instance, studies like LLMEval-Med (Zhang et al., 2025) and ER-Reason (Mehandru et al., 2025) present benchmarks grounded in real-world clinical workflows, using questions generated or reviewed by physicians. Others, such as the MedQA (D. Jin et al., 2020) and MedMCQA (Pal et al., 2022) benchmarks, leverage existing multiple-choice exams from the US and India respectively to assess reasoning and factual recall in a controlled setting.

## Datasets used in Quantitative Medical Reasoning Benchmarks

Medical reasoning benchmarks have been traditionally composed of question-answer datasets(Huang et al., 2024). Closed-ended questions have still retained relevance because of the relatively objective nature compared to subjective open ended questions.

AfriMed-QA is a regional benchmark designed to reflect African clinical realities by drawing exclusively from medical case content collected across 60 institutions in 16 African countries(Olatunji et al., 2024). It earned a distinct place for local context relevance and question diversity, incorporating multiple formats such as multiple-choice, short-answer, and consumer health questions. However, the benchmark lacks formal alignment with national or WHO guidelines, limiting its utility for evidence-based model evaluation. While its expert validation process was extensive—featuring 379 raters including 58 clinicians—the dataset is static, with no defined update mechanism. With over 15,000 questions and coverage across 32 specialties, AfriMed-QA represents a meaningful contribution to African medical NLP research, though uneven topic distribution and English-only support constrain its representativeness.

MedMCQA is a large-scale benchmark derived from Indian medical entrance exams (AIIMS and NEET PG), offering approximately 194,000 multiple-choice questions(Pal et al., 2022). Though the dataset is primarily focused on academic assessments in an LMIC context, it does not reflect African-specific healthcare scenarios. It similarly lacks guideline alignment, relying on exam syllabi rather than formal clinical protocols. Academic professionals carried out expert validation, though its one-time release in 2022 with no update path limits adaptability. Topic coverage is notably broad, spanning over 2,400 subtopics, making it valuable for assessing breadth in model understanding. Despite this, its question diversity is narrow, consisting only of MCQs, and it remains English-only.

MedQA-USMLE is based on the United States Medical Licensing Examination and includes over 60,000 items curated by professional exam writers(D. Jin et al., 2020). The dataset benefits from high topical breadth and a reasonable degree of expert validation, making it a reliable benchmark for evaluating knowledge-intensive reasoning. However, its relevance to low-resource settings is limited, as it is rooted in high-income, resource-rich clinical contexts. Additionally, while the USMLE itself is periodically updated, the MedQA-USMLE dataset is a static snapshot and lacks an update mechanism to reflect contemporary clinical knowledge. It includes English and Chinese versions, offering modest multilingual support, but does not account for the cultural or epidemiological context of African or tropical medicine.

PubMedQA is a unique benchmark focused on biomedical research comprehension, with over 273,000 QA pairs derived from PubMed abstracts(Q. Jin et al., 2019). While it offers high corpus volume and coverage across research domains such as oncology and immunology, it scores poorly on local context and clinical guideline alignment. The dataset's QA pairs are predominantly yes/no/maybe formats with long-form evidence, based on study conclusions rather than clinical guidance. Its expert validation is limited to a small subset (1,000 labeled questions), with the rest being unlabeled or AI generated. Moreover, it has not been updated since its 2019 release and its strict reliance on English-language biomedical literature limits its applicability to multilingual or context-sensitive clinical evaluation tasks.

## 2 Methodology

### 2.1 Datasets

We conducted a structured literature search across both peer-reviewed and grey literature sources, including PubMed, IEEE Xplore, ACM Digital Library, Scopus, arXiv, bioRxiv, and Google Scholar. Search terms focused on large language models (LLMs), benchmarking, datasets and evaluation metrics, with results filtered to retain only studies within the health or clinical domain that reported quantitative performance metrics. Qualitative-only papers such as commentaries and conceptual frameworks were excluded.

The Alama Health QA dataset pipeline leveraged retrieval-augmented generation (RAG) to produce regulator-compliant question-answer pairs grounded in authoritative sources from the Kenyan Ministry of Health. The process involved constructing a comprehensive digital knowledge base from the Clinical Guidelines for management and referral of common conditions at level 2 and 3(Dispensaries and Health Centres). This was intentionally chosen to ground the questions to the Kenyan context as well as to ensure the appropriate level of care. The paragraphs were chunked and indexed documents for efficient retrieval. Using RAG, an LLM was prompted to generate case based questions based on retrieved guideline chunks, along with four answer options and explanations referencing the source. This iterative process covered various guideline topics, ensuring diverse question formats and anchoring each question to verifiable text. The three models tested in the generative AI step were GPT 4o-mini, Gemini flash 2.0 lite and llama 3.1 with Gemini being used for the final pipeline.

### 2.2 Quantitative semantic analysis

Questions were lower-cased and stripped of punctuation. We tokenized using spaCy's en_ner_bc5cdr_md model, retaining sentence boundaries(Neumann et al., 2019). We ran spaCy Named Entity Recognition(NER) over each corpus, extracting medical terms. For each unique medical term string we counted total instances. NER outputs were exported to CSV for subsequent analysis. To visualize semantic structure, we embedded each entity string by averaging subword embeddings from using Word2Vec(Mikolov et al., 2013) and the BioBERT model(Lee et al., 2020). We then applied t-SNE to project into 2 D. For each dataset we plotted a random sample of 200 points and manually annotated representative labels to inspect cluster coherence. We also applied the semantic metrics in table 1

*Table 1: Quantitative semantic metrics*

| Metric | Formal Definition | Rationale / Interpretation |
|---|---|---|
| Recency | $\frac{\#rows\ containing > 1\ post-2020\ term}{N}$ | Freshness wrt modern guidelines (COVID-19, DTG). |
| Neglected Tropical Disease (NTD)score | $\frac{\#rows\ containing\ any\ NTD\ term}{N}$ | Mention of any of the 20 WHO defined neglected tropical diseases |
| Average Word Length | $AWL = \frac{\sum_{j=1}^{T} len(w_j)}{T}$ where T is the number of alphabetic tokens. | Technical term proxy; jargon → longer words. |
| Flesch-Kincaid Grade Level | An estimate of the school grade level needed to understand a piece of text. | A score of 7.0, for example, suggests that the text is suitable for a seventh-grade student. |
| Type-Token Ratio (TTR) | $\frac{\#of\ unique\ words}{Total\ number\ of\ words}$ | Measures the proportion of unique words in a text. A higher TTR indicates greater lexical diversity. |
| Measure of Textual Lexical Diversity (MTLD) | Mean length of sequential word strings (in tokens) that maintain a TTR above the 0.72 threshold | Reflects consistency of lexical diversity |
| Hyper-Dispersed Diversity (HDD) | The likelihood that each word type appears at least once in a sample. | Comparing lexical richness across different-sized texts |

## 2.2 Qualitative analysis

To review the content of the different benchmarks, a blinded, reproducible, data-engineering pipeline was implemented in Python to construct the evaluation corpus and the electronic rating instrument. Stage 1 randomly harmonised six heterogeneous question-answer benchmarks from AfriMed-QA, PubMedQA, MedMCQA, MedQA-USMLE, MMLU-Medical and the in-house Alama-Health-QA—into a single master dataframe. All non-essential fields are collapsed into a cleaned unified column. Random sampling of 25 questions per dataset was done with a fixed seed to create five reviewer pack; each pack always contained a Alama subset plus one rotating external dataset. The tool was then converted into a Kobocollect questionnaire. Every rating field was defined as required on a five-point Likert scale across five quality dimensions (clinical relevance, guideline alignment, clarity/completeness, distractor plausibility, and language/cultural fit) plus free-text comments. Logic expressions ensured that individual reviewers only saw items assigned to their ID, preserving the double-blind design and mitigating confirmation bias.

Reviewers also evaluated the process behind the curation of the datasets including the data source transparency, expert involvement and diversity, diversity of the question types, the updating mechanisms and the validation pipelines (table 2).

*Table 2: Qualitative methodology metrics*

| Metric Category | Specific Metric | Description / Evaluation Criteria |
|---|---|---|
| 1. Dataset Source Transparency | Source Documentation | Are original documents or datasets clearly cited and accessible? |
| | Source Credibility | Are sources authoritative (e.g., peer-reviewed, WHO, CDC, government health agencies)? |
| | Licensing & Permissions | Are data use rights and licensing terms clearly stated? |
| | Temporal Relevance | Are sources current or regularly updated to reflect evolving medical knowledge? |
| 2. Expert Involvement & Diversity | Number of Experts | How many experts were involved in dataset/question curation? |
| | Type of Experts | Were domain-specific experts (clinicians, pharmacists, epidemiologists) involved? |
| | Geographic Diversity | Were experts drawn from multiple regions or countries including Africa? |

|  | Institutional Diversity | Were experts affiliated with varied institutions (academic, public health, private, NGO)? |
|---|---|---|
|  | Demographic Diversity | Does the team reflect diversity in gender, race, and other identity markers? |
| **3. Question/Task Diversity** | Question Type Variety | Are multiple question types (MCQs, fill-in-the-blank, reasoning, diagnosis, etc.) represented? |
|  | Clinical Task Representation | Does the benchmark span various medical domains (e.g., diagnosis, treatment, prognosis, etc.)? |
|  | Difficulty Stratification | Are questions stratified by difficulty (easy, moderate, advanced clinical reasoning)? |
| **4. Updating Mechanism** | Update Frequency | How often is the dataset updated or reviewed for relevance and accuracy? |
|  | Version Control | Are dataset versions and changes transparently tracked? |
|  | Responsiveness to New Evidence | Is there a process to incorporate emerging guidelines or medical literature? |
| **5. Validation Pipeline** | Internal Validation | Was the dataset reviewed internally for consistency and correctness? |
|  | External Validation | Was an external review or benchmarking performed? |
|  | Inter-rater Agreement | Is agreement among annotators or experts measured and reported? |
|  | Error Analysis Process | Is there a systematic process to identify and fix mislabeled or ambiguous examples? |

# 3 Results

The literature search revealed 31 papers on quantitative benchmarking of LLMS in medicine with 19 unique english medical benchmark datasets. For purposes of analysis, we selected the 5 most common datasets used in published medical LLM papers. These were AfriMed-QA, MMLU-Medical, PubMedQA, MedMCQA, MedQA-USMLE.

## Semantic Analysis

The semantic analysis compared six medical QA datasets in terms of their size, linguistic complexity, and vocabulary richness(table 3). AfriMed-QA, with 15,275 rows, featured brief sentences (15 tokens) and moderate depth (4.7), making it relatively accessible with a Flesch-Kincaid Grade Level (FKGL) of approximately 11, and moderate lexical diversity (MTLD 34). PubMedQA, although similar in sentence length, showed deeper syntactic structures (5.8) and a higher FKGL of 16, indicating more complex scholarly language, coupled with high lexical variation (TTR 0.96, MTLD 31). MedMCQA presented the most compact and readable language (13 tokens, shallow trees, FKGL 10) but also the lowest lexical richness (MTLD 22). In contrast, MedQA-USMLE stood out for its extremely long, vignette-style questions (103 tokens, depth 10.6, FKGL 50), and exceptionally rich vocabulary (MTLD 94). MMLU-Medical combined concise yet information-dense content (55 tokens, depth 7.2, FKGL ~27) with strong vocabulary richness (MTLD 55). Finally, Alama-QA struck a balance with medium-long sentences (43 tokens, depth 7.9, FKGL 25) and ranked second in lexical richness (MTLD 75), making it both complex and diverse linguistically.

*Table 3: Semantic analysis results*

| **Dataset** | **Rows (n_rows)** | **"How hard is the language?"** *(Sentence length → Depth → Readability)* | **"How rich is the vocab?"** *(TTR → MTLD → HDD)* |
|---|---|---|---|
| AfriMed-QA | 15,275 | Brief (15 tokens, depth 4.7, FKGL ≈ 11) | Medium diversity (TTR 0.95, MTLD 34) |
| PubMedQA | 273,518 | Scholarly (15 tokens but deeper trees 5.8; FKGL ≈ 16) | Highest lexical variation (TTR 0.96, MTLD 31) |
| MedMCQA | 193,155 | Compact (13 tokens, shallow, FKGL ≈ 10) | Smallest MTLD (22) |
| MedQA-USMLE | 11,451 | Vignette-heavy (103 tokens, depth 10.6, FKGL ≈ 50) | MTLD high(94) |
| MMLU-Medical | 607 | Dense facts (55 tokens, depth 7.2, FKGL ≈ 27) | Strong diversity (MTLD 55) |
| Alama-QA | 40,607 | Mid-long (43 tokens, depth 7.9, FKGL ≈ 25) | Second-richest vocab (MTLD 75) |

# African Disease Burden Comparison between the benchmarks

Across all six benchmarks, Alama-QA emerged as the most representative of African health realities: it contained > 40 % of the total NTD-related term mentions and the highest within-set prevalence for every priority disease tracked (e.g., malaria 7.74 %, HIV 4.07 %, TB 5.15 %). AfriMed-QA was an important second, with reasonable coverage of malaria (1.51 %), sickle-cell disease (1.06 %), TB (1.07 %), and Ebola/VHFs (0.26 %). By contrast, the mainstream global benchmarks like MMLU-Medical, MedQA-USMLE, MedMCQA, and PubMedQA collectively contributed < 20 % of total NTD mentions and showed near-zero incidence for several critical conditions (e.g., malaria absent from MMLU-Medical, dengue absent from all four). Even when these datasets did reference high-burden African diseases, their proportions were minuscule (e.g., HIV 0.42 % in MedMCQA; malaria 0.19 % in MedMCQA; TB 0.25 % in PubMedQA). Overall, Alama-QA and, to a lesser extent, AfriMed-QA demonstrate far superior representativeness for diseases that disproportionately affect Africans, while the widely used international benchmarks remain heavily skewed toward pathologies common in high-income settings(table 4).

*Table 4: Count of benchmark questions on selected conditions with higher african burden*

| Term | Alama-QA | MMLU-Medical | MedQA-USMLE | MedMCQA | AfriMed-QA | PubMedQA |
|---|---|---|---|---|---|---|
| Dengue | 356 | 0 | 0 | 0 | 49 | 117 |
| Ebola / VHFs | 8 | 0 | 0 | 8 | 39 | 13 |
| Fever | 5 726 | 34 | 1 228 | 2 188 | 388 | 215 |
| HIV | 1 651 | 5 | 176 | 811 | 94 | 1 167 |
| Leishmaniasis | 23 | 0 | 0 | 0 | 37 | 44 |
| Malaria | 3 143 | 0 | 13 | 362 | 230 | 368 |
| Pneumonia | 3 032 | 7 | 133 | 429 | 131 | 530 |
| Sickle-cell disease | 299 | 0 | 0 | 0 | 162 | 165 |
| Tuberculosis | 2 091 | 0 | 48 | 529 | 163 | 647 |

We analyzed expert reviewer feedback across six prominent medical QA datasets: Alama-Health-QA, MedMCQA, MedQA-USMLE, AfriMed-QA, MMLU-Medical, and PubMedQA. Using thematic analysis, we identified five key domains that captured reviewer perceptions: (1) Clinical Relevance, (2) Guideline Alignment, (3) Clarity & Completeness, (4) Plausibility of Distractors, and (5) Language & Cultural Appropriateness(figure 2).

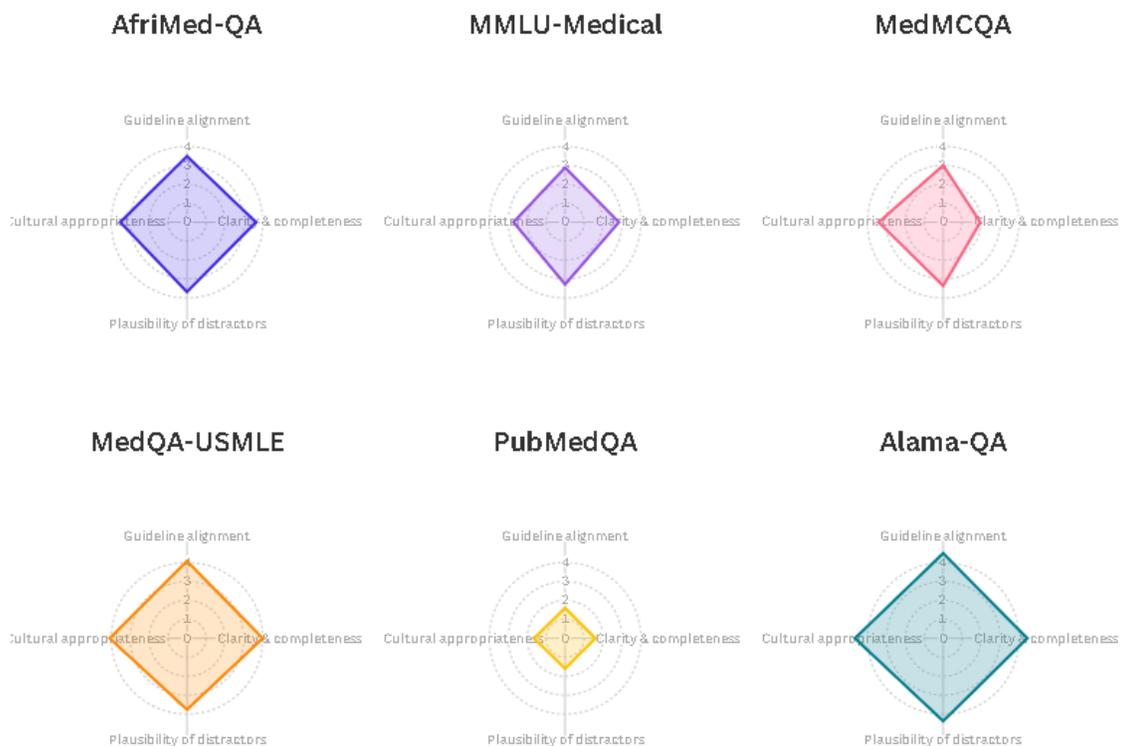

Figure 2: Radar graphs of qualitative evaluation of various benchmark datasets

# Qualitative evaluation

### 1. Clinical Relevance and Guideline Alignment

**AfriMed-QA** with a score of 3.9/5 was noted to have some questions with reference to existing medical guidelines e.g. Diagnostic and Statistical Manual of Mental Disorders(DSM-5) which was a huge plus for reviewers.

> *Question: According to DSM 5, for a diagnosis of Bipolar 1, at least one episode of which the following is required? answer_options: "Depressive", "Manic", "Hypomanic","Dysthymic", "B and C" correct_answer: option2*
> *Reviewer comment: "Good case with mention of DSM 5'*

**MMLU-Medical** scored 3.5/5 and attempted to simulate clinical scenarios, such as surgical aftercare, but received mixed reviews due to a lack of context specificity.

> *Question: A patient has been on the operating table for four hours. How long may it take for any pressure damage to be visible?*
> *Options: A) 12 hours, B) 72 hours, C) 24 hours, D) 5 days*
> *Correct Answer: A*
> *Reviewer Comment: "Pressure damage may not be as visible on black skin. It may take longer for it to be visible and so the answers should also take this into account."*

**PubMedQA** with a score of 2.2/5 was widely critiqued for lacking clinical framing altogether. Questions were often derived from research literature without being transformed into decision-support or diagnostic items.

> *Question: Is endothelin-1 an aggravating factor in the development of acute pancreatitis?*
> *Final Answer: Yes*
> *Reviewer Comment: "Too academic."*

Reviewers consistently rated **Alama Health-QA** items as highly clinically relevant (4.7/5), especially those simulating scenarios faced by frontline health workers. Questions related to child malnutrition, HIV, and community-level decision-making were viewed as rooted in real Kenyan practice.

> *Question: A community health volunteer is conducting a home visit in Kibera. She finds a family with several children, one of whom is severely malnourished with visible signs of edema. The family struggles to afford food. What is the most appropriate immediate action?*
> *Options: A) Provide the family with a prescription for antibiotics. B) Advise the family to increase their intake of carbohydrates. C) Refer the child to the nearest health facility for assessment and management. D) Tell the family to give the child more water.*
> *Correct Answer: C*
> *Reviewer Comment: "Great question, aligns with guidelines."*

By contrast, reviewers found **MedMCQA** (3.7/5) to be clinically relevant in subject matter but often diluted by unnecessary medical detail. One example concerning platelet storage was considered textbook-relevant but poorly tailored for clinical application.

## 2. Clarity and Cognitive Load

Clarity of language and avoidance of cognitive overload are crucial for usability and educational value. While **Alama Health-QA** scored 4.5/5 with language that was accessible to early-career clinicians or community health workers, reviewers found some technically light questions.

> *Question: A healthcare worker is providing care to a patient who is a PWID. Question: What is the recommended first-line ART for adult PWID? A) TDF + 3TC + DTG B) AZT + 3TC + EFV C) TDF + FTC + LPV/r D) ABC + 3TC + DTG*
> *Correct: A*
> *Reviewer comments: "It would help to have additional information e.g. if the patient is on any ART medication and the parameters."*

In contrast, **MedMCQA** with a score of 3/5 frequently overwhelmed reviewers with verbose explanations and extended rationales.

> *Question: Intercalated discs are seen in   a: Iris   b: Cardiac muscle   c: Musculotendinous endings   d: Nerve bundles   exp: Ans: b) Cardiac muscleIntercalated discs are cell membranes that separate individual cardiac muscle cells from one another. Cardiac muscle fibres are made up of many individual cells connected in series and in parallel in such a way that they form permeable "communicating junctions"- gap junctions that allow free diffusion of ions. ?*
> *Reviewer Comment: "Good question but answer is too long."*

**AfriMed-QA** also suffered from overgeneralized or ambiguous phrasing, which sometimes obscured intent. While the effort to include cultural relevance was appreciated, vague terminology such as "some cultures" undermined the question's specificity and clarity.

## 3. Plausibility of Distractors

Reviewers valued datasets that featured answer options closely resembling plausible clinical errors or misconceptions. While **Alama Health-QA** scored highly here with 4.4/5, some distractors were too obvious to the seasoned clinician.

> *Question: A 50-year-old man from a rural area presents with severe abdominal pain, vomiting, and altered mental status. He admits to taking an excessive amount of a common pain reliever. Question: Based on the provided information, which of the following is the MOST likely cause of his symptoms? A) Iron tablet overdose B) Digoxin toxicity C) Paracetamol overdose D) Opiate overdose Correct: C*
>
> *Reviewer Comment: "Distractors were too obvious as none of them are "common " painkillers."*

**MedMCQA**, while often containing technically accurate distractors, scored lower at 3.4/5 and was criticized for having too many abbreviations. One reviewer noted that the use of abbreviations interfered with the quality of the distractor options.

> *Question: Which of the following is not a content of plane LA   a: Methyl paraben  b: Epinephrine  c: Nacl  d: Lignocaine*
>
> *Reviewer Comment: "Unclear question. Use of abbreviations. "*

### 4. Language and Cultural Appropriateness

The extent to which questions respected local context and patient diversity was a key concern. **AfriMed-QA** explicitly attempted to reflect African traditions and beliefs, which reviewers appreciated with a high score of 4.5/5. However, some items were too vague, reducing cultural nuance to generalized statements.

> *Question: How do certain African societies manage symptoms of earwax buildup?*
> *Answer Rationale: Instilling a few drops of warm olive oil or hydrogen peroxide into the ear canal is a traditional method used to soften earwax and facilitate its removal.*
> *Reviewer Comment: "Only for some cultures, it is a bit too wide of a net to narrow down on hydrogen peroxide and olive oil."*

**PubMedQA** was perceived as too detached from any meaningful cultural framing. The academic tone and research-like phrasing alienated it from real-world clinical discourse in low-resource or African settings reflecting its poor score of 1.6/5.

# Discussion

This study highlights a significant representational gap between widely used medical LLM benchmarks and the disease burdens and clinical realities of African health systems mirroring existing work on algorithmic bias(Ukanwa, 2024). While existing global benchmarks like MedQA-USMLE, PubMedQA, and MedMCQA are well-established in evaluating factual recall and reasoning performance of LLMs, their geographic, linguistic, and epidemiological biases limit their utility in African contexts. Our findings suggest that overreliance on these benchmarks risks propagating both epistemic and operational harms: epistemic, by reinforcing a narrow model of global medicine skewed toward high-income diseases and care paradigms; and operational, by undermining regulatory alignment, safety, and clinical utility in African health settings. This is in line with already established differences in Western and African medical knowledge systems(Asakitikpi, 2018).

In this work we present Alama Health QA, a pipeline for creating contextual and regulator-specific benchmarks for Africa(Mutisya et al., 2025-in print). Alama-Health-QA was developed using retrieval-augmented evaluation anchored on the Kenyan Ministry of Health guidelines. This concept of using Agentic AI to augment an existing knowledge base into a dataset has been tried and tested successfully(L. Liu et al., 2024). However, this is the first instance where Ministry of Health validated knowledge bases have been used for the benchmark generation. Because of this grounding, it demonstrated strong alignment with local treatment protocols, high lexical diversity, and superior cultural and linguistic contextualization. It was also rated highest in clinical relevance and plausibility of distractors. This supports the argument that benchmark construction grounded in local guidelines and health worker scenarios can substantially improve the validity of LLM evaluations for low-resource settings.

Notably, Alama-Health-QA and AfriMed-QA outperformed global counterparts in capturing diseases with disproportionate African burden—such as malaria, tuberculosis, HIV, and sickle-cell disease. Sickle cell disease, a condition with over 4 million affected individuals and particularly prevalent in tropical regions, serves as a striking example of underrepresentation—receiving no mentions at all in three of the benchmark datasets(Adigwe et al., 2023).

Conversely, global benchmarks demonstrated strong technical merits—such as vocabulary richness and complexity in MedQA-USMLE, and breadth of subtopics in MedMCQA—but were weak on contextual relevance and disease coverage. For instance, several high-burden African conditions, including dengue and leishmaniasis, were nearly absent from the mainstream datasets. Additionally, qualitative assessments showed that datasets like PubMedQA and MMLU-Medical offered limited clinical decision support value due to either excessive abstraction or overreliance on research language rather than frontline scenarios.

The thematic review further emphasized the disconnect between benchmark quality and clinical usability. Even well-validated datasets such as MedMCQA and MMLU-Medical suffer from ambiguity, cultural insensitivity, or verbosity, which hinders their applicability for training or evaluating models intended for diverse populations. This disconnect was particularly evident in guideline alignment, where only Alama-Health-QA embedded national treatment protocols. The lack of update mechanisms in most datasets also presents a critical limitation, especially in fast-evolving areas like HIV care and pandemic preparedness.

Importantly, while AfriMed-QA made a laudable effort to reflect African clinical scenarios, it suffered from an absence of formal linkage to WHO or national guidelines, an uneven topic distribution, and a static release format. Nevertheless, its inclusion of multiple question formats and pan-African data sources positions it as a valuable complementary resource.

Together, these findings support the need for regionalized, regulator-aligned, and linguistically diverse benchmarks that go beyond transplanting exam syllabi from high-income countries. Benchmarks like Alama-Health-QA serve as a proof-of-concept that local contexts, when properly digitized and structured, can support robust model evaluation and safer deployment. Future benchmarks should prioritize update cadence, multilingual support, and alignment with disease surveillance priorities across LMICs. Furthermore, funders and research communities should invest in capacity-building efforts to enable African institutions to co-create and govern benchmark standards that reflect their own health needs and values.

## Limitations

This study focused on English-language QA benchmarks and did not utilize the swahili iteration of Alama Health QA for the purposes of fair comparison. We also excluded open-ended or generation-focused datasets like Health Bench since we have a separate comparison paper in development. Furthermore, although qualitative reviews were carefully blinded and standardized, they remain subject to human interpretive variability. Lastly, this review did not assess model performance directly but instead evaluated benchmark structure and representativeness—work is ongoing to triangulate benchmark quality with LLM output behavior in African clinical scenarios.

# Conclusion

This study makes a compelling case that benchmark representativeness is not merely a technical nicety but a clinical and ethical necessity. The dominance of benchmarks rooted in high-income clinical contexts limits the utility, safety, and fairness of LLMs for low- and middle-income countries (LMICs), particularly in Africa. Regional benchmarks such as Alama Health-QA show that it is feasible—and essential—to design evaluation tools that reflect national guidelines, local diseases, and the real-world conditions of frontline healthcare.

# Next Steps

The next step is to develop and release disease-specific benchmarks tailored to underrepresented yet high-burden conditions in Africa. Based on our findings, HIV, Malaria, Sickle-cell disease, and viral hemorrhagic fevers (e.g., Ebola) are top candidates for such condition-focused QA datasets. These benchmarks should build on the methodology of Alama Health-QA, ensuring linkage to authoritative guidelines, inclusion of culturally and linguistically appropriate phrasing, and stratified clinical reasoning difficulty.

For readers interested in replicating or extending this work, the methodological framework for Alama-Health-QA—detailing the RAG pipeline, data sources, prompt design, and quality control—is described in the accompanying methodological paper titled "Alama-Health-QA: A Guideline-Grounded Benchmark Pipeline for Creating Medical Language Models Benchmarks in African Primary Care." This paper outlines how local clinical knowledge can be operationalized into scalable, verifiable LLM evaluation tools for public sector use.

# Appendix- Additional Notes on the Benchmark Methodology

## Summary Score of Benchmark Methodology

| Metric | AfriMed | MedMCQA. | MedQA-USMLE | PubMedQA | Alama Health QA |
|---|---|---|---|---|---|
| Local Context Relevance | ★★★★★ | ★★ | ★ | ★ | ★★★★★ |
| Guideline Alignment | ★ | ★ | ★ | ★ | ★★★★★ |
| Expert Validation | ★★★★ | ★★ | ★★★ | ★★ | ★★ |
| Update Mechanism | ★ | ★ | ★★★ | ★★ | ★★★★ |
| Corpus Size | ★★★ | ★★★★★ | ★★★ | ★★★★★ | ★★★★★ |
| Topic Coverage | ★★★ | ★★★★★ | ★★★★ | ★★★★ | ★★★★ |
| Question Diversity | ★★★★ | ★★★ | ★★★ | ★★★ | ★★★ |
| Multilingual Support | ★ | ★ | ★★ | ★ | ★★ |

## AfriMed-QA

| Metric | Score | Reviewer Notes |
|---|---|---|
| Local Context Relevance | ★★★★★ | Sourced entirely from African clinical scenarios via 60 schools in 16 countries. |
| Guideline Alignment | ★ | No explicit mapping to national or WHO clinical guidelines described. |
| Expert Validation | ★★★★ | Human evaluation involved 379 raters, including 58 clinicians. |
| Update Mechanism | ★ | One-off release in 2024; no formal cadence for refreshing with new guidelines. |
| Corpus Size | ★★★ | 15,275 total Qs—smaller than global benchmarks but substantial for a regional dataset. |
| Topic Coverage | ★★★ | Covers 32 specialties. Unequal topic sizes skews its performance to key topics like maternal health while topics like rheumatology are neglected |
| Question Diversity | ★★★★ | Includes MCQs, SAQs, yes/no, factoids, and consumer queries. |
| Multilingual Support | ★ | English only (plans to add other African languages in future work). |

## MedMCQA

| Metric | Score | Reviewer Notes |
|---|---|---|
| Local Context Relevance | ★★ | Focused on Indian PMET & NEET PG entrance exams—LMIC relevance but not African. |
| Guideline Alignment | ★ | Based on exam syllabi, not on national or WHO care protocols. |
| Expert Validation | ★★ | MCQs sourced from high-stakes AIIMS & NEET PG exam banks and validated by education experts. |
| Update Mechanism | ★ | Single release in 2022; no formal update cycle. |
| Corpus Size | ★★★★★ | ~194 k MCQs covering extensive exam content. |
| Topic Coverage | ★★★★★ | Spans 2.4 k topics across 21 medical subjects. |
| Question Diversity | ★★★ | MCQs with detailed rationales; lacks open-ended or dialog formats. |
| Multilingual Support | ★ | English only. |

## MedQA-USMLE

| Metric | Score | Reviewer Notes |
|---|---|---|
| **Local Context Relevance** | ★ | USMLE vignettes; minimal tropical/resource-limited content. |
| **Guideline Alignment** | ★ | Aligned to U.S. exam standards, not to Country Specific/WHO protocols. |
| **Expert Validation** | ★★★ | Authored and vetted by professional board exam writers. |
| **Update Mechanism** | ★★★ | Developed in 2020. USMLE updates periodically, but dataset snapshot is static. |
| **Corpus Size** | ★★★ | 61,097 total Qs |
| **Topic Coverage** | ★★★★ | Broad clinical specialties as covered in board exams. |
| **Question Diversity** | ★★★ | Primarily MCQs; few open-ended items. |
| **Multilingual Support** | ★★ | English, Simplified Chinese, and Traditional Chinese. |

## PubMedQA

| Metric | Score | Reviewer Notes |
|---|---|---|
| **Local Context Relevance** | ★ | Based solely on PubMed research abstracts—not clinical scenarios. |
| **Guideline Alignment** | ★ | Answers derived from study conclusions, not linked to care guidelines. |
| **Expert Validation** | ★★ | 1 k expert-annotated instances; large unlabeled and generated subsets. |
| **Update Mechanism** | ★★ | Static 2019 snapshot; no ongoing refresh described. |
| **Corpus Size** | ★★★★★ | ~273 k total QA instances (1 k labeled + 61.2 k unlabeled + 211.3 k generated). |
| **Topic Coverage** | ★★★★ | Broad across biomedical research domains (oncology, immunology, mechanistic studies). |
| **Question Diversity** | ★★★ | Primarily yes/no/maybe format with long-answer conclusions. |
| **Multilingual Support** | ★ | English abstracts only. |

## Alama Health QA

| Metric | Score | Reviewer Notes |
|---|---|---|
| **Local Context Relevance** | ★★★★★ | Location metadata in the questions with references of african places and personas |
| **Guideline Alignment** | ★★★★★ | Directly tied to guideline metadata of the Kenya Clinical Management Guidelines |
| **Expert Validation** | ★★ | Expansion to wider multidisciplinary validation team planned |
| **Update Mechanism** | ★★★★ | Tied to guideline update schedule which can be 2-3 years. Metadata allows targeted updates |
| **Corpus Size** | ★★★★★ | Dynamic based on guideline source |
| **Topic Coverage** | ★★★★ | Targeted |
| **Question Diversity** | ★★★ | Primarily Multiple choice case study format with long-answer conclusions. |
| **Multilingual Support** | ★★ | English with Swahili variant created |